\documentclass[10pt,twoside]{article}

\usepackage{times}
\usepackage[utf8]{inputenc}
\usepackage[T1]{fontenc}
\usepackage{graphicx}

\usepackage{taln2017}
\usepackage[frenchb]{babel}

\title{Analyse automatique FrameNet : une étude sur un corpus français de textes encyclopédiques}

\author{Gabriel Marzinotto\up{1, 2}\quad Géraldine Damnati\up{1}\quad Frédéric Béchet\up{2}\\
  {\small
    (1) Orange Labs, Lannion, France \\ 
    (2) Aix-Marseille Université, Marseille, France \\ 
    \texttt{\{gabriel.marzinotto,geraldine.damnati\}@orange.com\quad frederic.bechet@lif.univ-mrs.fr \\
  }}}

\begin{document}
\maketitle

\resume{
Cet article présente un système d'analyse automatique en cadres sémantiques évalué sur un corpus de textes encyclopédiques d'histoire annotés selon le formalisme FrameNet. L'approche choisie repose sur un modèle intégré d'étiquetage de séquence qui optimise conjointement l'identification des cadres, la segmentation et l'identification des rôles sémantiques associés. Nous cherchons dans cette étude à analyser la complexité de la tâche selon plusieurs dimensions. Une analyse détaillée des performances du système est ainsi proposée, à la fois selon l'angle des paramètres du modèle et de la nature des données.}

\abstract{FrameNet automatic analysis: a study on a French corpus of encyclopedic texts}{
  This article presents an automatic frame analysis system evaluated on a corpus of French encyclopedic history texts annotated according to the FrameNet formalism. The chosen approach relies on an integrated sequence labeling model which jointly optimizes frame identification and semantic role segmentation and identification. The purpose of this study is to analyze the task complexity from several dimensions. Hence we provide detailed evaluations from a feature selection point of view and from the data point of view.
}

\motsClefs
  {Analyse en cadres sémantiques, étiquetage de séquence, textes encyclopédiques}
  {Semantic frame analysis, sequence labeling, encyclopedic texts}

\section{Introduction}
\vspace{-5pt}
L'extraction d'informations structurées dans des textes est un préalable qui favorise l'accès aux connaissances qu'ils contiennent par des outils de Traitement Automatique du Langage. Dans cette étude, nous nous intéressons au cas particulier de textes encyclopédiques historiques et nous nous plaçons dans le contexte de la représentation sémantique FrameNet. Dans l’approche FrameNet initiée par l’Institut ICSI  de Berkeley \cite{Baker:1998:BFP:980845.980860}, un cadre sémantique (\textit{Frame}) peut être évoqué par des unités lexicales (les triggers ou cibles). Par exemple, le cadre "Commitment" peut être évoqué par "promettre", "promesse", "s’engager" et le cadre "Becoming\_aware" peut être déclenché par "découvrir" et "découverte". Les unités lexicales (UL) qui déclenchent un cadre peuvent être nominales ou verbales. Par ailleurs, un cadre englobe la définition des participants et des propriétés qui peuvent lui être attachés : ce sont les \textit{Frame Elements} (FE). Ils sont spécifiques à chaque cadre et sont nommés par des labels explicites. Par exemple, dans la phrase suivante, l’agent de l’action de découvrir, est représenté par le FE "Cognizer" qui a deux instances :

\hspace{1cm}[le premier Européen]\texttt{Cognizer} à avoir [découvert]\texttt{Becoming\_aware} 

\vspace{-0.2cm}\hspace{1cm}[Mammoth Cave]\texttt{Phenomenon} était [John Houchin]\texttt{Cognizer}, [en 1797]\texttt{Time}.

Les cadres peuvent être liés entre eux par des relations \cite{fillmore2004framenet} (ex : inheritence, using, …) auquel cas les FE peuvent être mis en correspondance. Dans cette étude, nous réalisons une analyse "à plat" sans mettre les cadres en relation. 
Si les ressources linguistiques décrivant ces cadres sont de plus en plus nombreuses pour la langue anglaise, leur constitution pour le français n’en est qu’au début avec  les contributions du projet ASFALDA qui s’est attaché à produire des  ressources sur la base de FrameNet pour le français \cite{djemaa-etal-2016}. Pour notre part, nous avons constitué le corpus CALOR \citet{calorframe-2017} annoté en cadres sémantiques sur des textes encyclopédiques issus de différentes sources, dans le domaine de l'histoire, décrit plus en détail à la section \ref{corpus}.

L’analyse en cadres sémantiques a pour objectif de repérer dans des documents des instances de cadres avec l’ensemble des rôles permettant de les caractériser, et se fait classiquement en deux étapes.
La première est une étape de désambiguïsation afin d’identifier un cadre étant donnée la présence d’un déclencheur potentiel (UL)  
La seconde consiste à identifier les rôles sémantiques (FE) et est le plus souvent traitée de façon séquentielle comme l'enchaînement d'une étape de détection de segment et de classification de ce segment \cite{johansson2012semantic,lechelleutilisation}.
Le système Semafor \cite{das2014frame} constitue à l’heure actuelle une référence dans le domaine. Dans Semafor, l’étape d’identification de cadre, étant donné un déclencheur, est réalisée à l’aide d’un classifieur probabiliste par Maximum d’Entropie. Ensuite, l’étape de labélisation des rôles sémantiques est réalisée à l’aide d’un modèle log-linéaire conditionnel qui catégorise des segments (labélisation des FE). Dans cette étape, les segments candidats sont obtenus à partir de l'analyse en dépendance et correspondent aux sous-arbres produits. De cette façon, le modèle arrive à gérer à la fois la classification et la segmentation. 

Dans cette étude, nous adoptons une approche plus intégrée où les étapes de désambiguïsation, de détection des FE et de labélisation des FE se font de façon simultanée à l'aide de modèles d'étiquetage de séquences de type Conditional Random Fields (CRF). Notre modélisation du problème n'impose pas la contrainte que les FE soient toujours la projection d'une tête dans l'arbre de dépendances, ce qui  rend le système robuste aux erreurs d’analyse en dépendance. Nous avons décidé de travailler avec les CRF car ce sont des modèles simples qui ne nécessitent pas de grandes puissances de calcul, ni de gros volumes de données en comparaison avec les méthodes neuronales, mais qui sont suffisamment performants pour nous permettre la mise en place de nos expériences contrastives.  

La section \ref{tache} présente en détail cette approche avec un focus sur le choix des paramètres des modèles. La section \ref{results} présente un ensemble d'expériences visant à montrer l'influence du choix des paramètres, l'impact de la complexité intrinsèque des textes, et l'influence des données d'apprentissage.
\vspace{-8pt}

\vspace{-0,3cm}
\section{\label{tache}Analyse en cadres comme une tâche d'étiquetage de séquence}
\vspace{-5pt}


Les CRF ont été utilisés dans de nombreuses tâches du TALN ainsi que les modèles neuronaux de type RNN ou LSTM \cite{hakkani2016multi,tafforeau2016joint}. Nous avons choisi dans cet article d'utiliser des modèles CRF en se focalisant sur des comparaisons entre différents systèmes de traits et différents corpus pour entraîner et tester nos systèmes. Nous nous intéressons également à évaluer les effets du choix du corpus d’apprentissage en considérant des textes de différents styles (encyclopédique, encyclopédique adressé aux enfants, etc.) et qui traitent de sujets différents (archéologie, histoire, etc.).  
Vu que l’apprentissage se fait sur des corpus de taille moyenne et que notre objectif est de faire une étude comparative de la tâche, et non pas d’arriver aux meilleures performances possibles du système final, nous avons décidé de travailler avec les modèles CRF, car ils sont plus simples, plus rapides en apprentissage et présentent moins de paramètres à régler. 

Apprendre un seul CRF avec tous les exemples de phrases annotées résulterait en un très grand nombre d’étiquettes, ce qui peut être rédhibitoire lorsqu’on augmente le nombre de cadres. Une autre possibilité est de modéliser chaque cadre (sens) avec un CRF, mais cela nous obligerait à mettre en place un modèle de désambiguïsation en amont, de manière à savoir quel est le CRF qui doit être appliqué à chaque unité lexicale. Pour éviter ces problèmes nous avons décidé de modéliser chaque UL avec un CRF, cela permet de faire en même temps la désambiguïsation de cadres, la détection et la sélection des rôles sémantiques. Ce choix n’est pas nécessairement optimal dans le sens où il disperse les données d’apprentissage et ne permet pas le partage d’information entre des UL qui se ressemblent. Néanmoins il permet de passer à l’échelle lorsqu’on augmente le nombre de cadres. 

Ainsi, pour analyser en cadres une nouvelle phrase, nous allons d’abord extraire les UL de la phrase qui apparaissent dans notre liste de 145 UL possibles. 
Pour chaque phrase il y aura autant d'applications de CRF qu'il y a d’UL, puis une étape de post-traitement permet de vérifier la cohérence des résultats d'étiquetages. Ici nous vérifions que les étiquettes mises sur les rôles sémantiques sont compatibles et rattachables aux types de cadres sémantiques prédits par les CRF. 
Il est possible en effet que le CRF prédise un rôle sémantique qui ne fasse pas partie des rôles possibles du cadre mais qui ferait partie des rôles d'un autre cadre qui pourrait être déclenché par la même UL. Dans notre modèle tous les rôles qui ne sont pas rattachables à leur cible sont systématiquement rejetés. 



Pour extraire des caractéristiques pertinentes à la tâche nous avons évalué plusieurs types de paramètres et de codages inspirés de la littérature \cite{das2014frame,michalon:hal-01391678}. 
Une sélection incrémentale a été faite pour ne retenir que les 5 paramètres les plus pertinents pour chaque token : 
\begin{itemize}
\item son lemme, le lemme du mot parent, sa partie du discours (POS), la distance linéaire à la cible et les deux derniers niveaux du chemin de dépendances entre le mot et la cible.
\end{itemize}

La distance linéaire à la cible est le nombre de tokens entre le token courant et l’UL qui déclenche le cadre (cible). Ce nombre est négatif si le token est avant la cible, ou positif s'il est après. 
Le chemin de dépendance vers la cible se construit comme la concaténation des dépendances entre le token courant et la cible. L'analyse syntaxique est réalisée à l'aide de l’analyseur MACAON \cite{macaon:2010} qui construit des arbres syntaxiques avec un jeu de dépendances très similaire à celui du French TreeBank \cite{abeille2003building,abeille2004enriching}. Dans le cas général la cible qui déclenche un cadre n’est pas nécessairement la racine de l’arbre de dépendance de la phrase, cela implique que le chemin de dépendances entre un token et une cible est composé des dépendances non pas seulement de fils à parent (relations ascendante), mais aussi de parent à fils (relations descendantes). Nous faisons cette distinction de manière explicite en codant les chemins ascendants et descendants avec des symboles différents. Par ailleurs, nous avons observé que les chemins syntaxiques très longs étaient difficiles à modéliser. Pour contourner ce problème nous avons étudié la simplification de ces chemins en limitant leur longueur maximale, c'est-à-dire, lorsque le chemin de dépendances d'un token vers la cible du cadre sémantique dépasse une certaine longueur, nous allons le représenter avec un chemin plus court qui garde la plus grand quantité d'information possible. Dans nos expériences, nous avons obtenu que la simplification qui produisait les meilleures performances consiste à garder les deux dépendances du chemin les plus proches de la cible, qui sont souvent les plus pertinents.

\vspace{-0,4cm}
\section{\label{results}Evaluation}
\vspace{-5pt}
\subsection{\label{corpus}Protocole expérimental}
\vspace{-5pt}
Nous avons réalisé toutes nos expériences sur le corpus CALOR. 
Il est constitué de documents issus de 4 sources différentes : le portail Wikipédia sur l’Archéologie (WA, 201 documents), le portail Wikipédia sur la Première Guerre Mondiale (WGM, 355 documents), des textes issus de Vikidia (VKH, 183 documents), l'encyclopédie en ligne pour enfants, à partir de deux portails (Préhistoire et Antiquité) et des textes historiques de ClioTexte (https://clio-texte.clionautes.org/) sur la Première Guerre Mondiale (CTGM, 16 documents). 
Annoter un corpus en cadres sémantiques n’est pas une tâche facile à aborder car le nombre de cadres et d’unités lexicales (UL) porteuses de sens que l’on pourrait définir est énorme. Dans le cas de FrameNet \cite{Baker:1998:BFP:980845.980860}, le dictionnaire des cadres sémantiques pour l’anglais, propose 1222 cadres possibles et 13615 UL à ce jour. Pour cette raison, un corpus annoté en cadres n’est souvent étiqueté que sur une sélection des cadres et UL les plus pertinents. Les UL en dehors de cette sélection restent sans annotation et une UL sélectionnée apparaissant dans un texte avec un sens qui n’est pas prévu dans notre dictionnaire de cadres sémantiques simplifié, nous lui attribuons un cadre spécial « OTHER ». 
Sur le corpus CALOR, 21.398 occurrences de cadres sémantiques ont été annotées, déclenchées par une des 145 UL présentes dans notre liste de UL traitables. 
Au total, 53 cadres sémantiques différents ont été annotés, auxquels s'ajoute le cadre OTHER. 

Lorsqu’une phrase est étiquetée en cadres sémantiques, il y a 4 sous-tâches qui se développent, parfois de façon implicite. Nous les avons incluses dans notre protocole car elles permettent d’évaluer très précisément les systèmes d’analyse en cadres sémantiques. Ce sont les tâches de : détection de cibles (DC) qui revient à décider si une UL doit être associée à OTHER ou non; sélection du bon cadre (SC) pour chaque cible détectée; détection des segments qui constituent des rôles sémantiques (DR); sélection des types de rôles sémantiques (SR). Même si l'ensemble de ces tâches est réalisé par un seul modèle intégré nous présentons les différents niveaux d'évaluation, avec un accent plus particulier sur le SR, sous-tâche qui est, de façon générale, la plus difficile de l'analyse en cadres sémantiques.

Le corpus a été divisé en cinq parties de sorte qu'aucun document ne soit jamais sous divisé et de sorte que la distribution des cadres soit la plus homogène possible entre chaque partie. Pour chaque expérience nous mesurons la précision, le rappel et la F-mesure moyennés entre les 5-Folds ainsi que l’écart type des mesures de performances sur les 5 folds. 

\vspace{-10pt}

\subsection{Évaluation globale et influence des paramètres}
\vspace{-5pt}
Dans le tableau \ref{table:BestSystem} nous montrons les performances du meilleur système développé à partir des 5 caractéristiques les plus pertinentes pour la tâche. Sur ce corpus les tâches DC et SC ont une complexité assez basse car nous traitons un nombre de cadres limité. 
Sur ces deux tâches, notre système CRF augmente la précision de 5 points par rapport à un système naïf qui choisirait la classe majoritaire. 
Comme ces sont des tâches simples dans notre corpus, nous arrivons à des performances élevées et assez proches car la proportion d'UL pouvant conduire à plusieurs cadres différents est assez faible (seulement 12 UL). La détermination de la catégorie OTHER demeure la principale difficulté à ce niveau. Par ailleurs, la tâche de SR qui est la plus complexe, présente un taux de précision acceptable (82.2\%) étant donné le nombre de rôles possibles (150 au total), mais les performances en termes de rappel sont à peine de 51.2\%. La performance élevée du système en termes de précision est due au fait d’avoir un modèle CRF pour chaque UL, car ceci diminue le nombre d’étiquettes (et le nombre de confusions) possibles au moment des prédictions. 

\begin{table}[h]
  \begin{center}
    \begin{tabular}{ c | c | c | c | }
      \cline{2-4}
       & Précision & Rappel & Fmesure \\ \hline
      \multicolumn{1}{ |c|  }{ Détection de Cible  (DC)} & $96.4 \pm 0.2$  & $96.4 \pm 0.2$  & $96.4 \pm 0.1$  \\ \hline
      \multicolumn{1}{ |c|  }{ Sélection de Cadre  (SC)}  & $ 95.3 \pm 0.4  $ & $ 95.2 \pm 0.2 $ & $ 95.3 \pm 0.2$ \\ \hline
      \multicolumn{1}{ |c|  }{ Détection des Roles (DR)}  & $ 89.7 \pm 0.5 $ & $ 55.9 \pm 0.7 $ & $ 68.8 \pm 0.5 $ \\ \hline
      \multicolumn{1}{ |c|  }{ Sélection des Roles (SR)}  & $ 82.2 \pm 0.6 $ & $ 51.2 \pm 0.7 $ & $ 63.1 \pm 0.6 $ \\
      \hline
    \end{tabular}
    \caption{évaluation par niveaux avec la meilleure configuration (CRF à 5 paramètres)}
    \label{table:BestSystem}
  \end{center}
\end{table}
\vspace{-12pt}
S'il est difficile de comparer avec les résultats obtenus par le système SEMAFOR (données en anglais en plus grande quantité, nombre de cadres modélisés supérieur,...), notons cependant que notre évaluation (SR) correspondrait à la tâche \textit{Argument Identification}, avec la configuration \textit{full parsing} car nous ne fixons pas de valeurs Oracle dans les étapes intermédiaires, et l'évaluation \textit{partial matching} car nous ne comptons pas les erreurs de frontière sur les rôles sémantiques. Dans ces conditions le meilleur système évalué dans \cite{das2014frame} conduit à une F-mesure de $50.24$. 

Pour chaque cadre, ses rôles sémantiques peuvent être interprétés comme des réponses à certaines questions que l’on peut poser sur le cadre. Par exemple, pour le cadre \textit{Deciding} nous avons : 

{\small \hspace{0,5cm} \texttt{Cognizer} (\textit{qui est l'agent?}) prend une \texttt{Decision}  (\textit{quoi?}) parmi \texttt{Possibilities} (\textit{parmi quoi?}) 

\vspace{-0.2cm}\hspace{0,5cm} parce que \texttt{Explanation} (\textit{pour quelle raison?}) à un \texttt{Time} (\textit{quand?}) et dans un \texttt{Place} (\textit{où?}).}

Ceci permet de regrouper les rôles sémantiques de différents cadres et de leur donner une interprétation simple qui aide à analyser quelles sont les questions génériques pour lesquelles notre système est capable de trouver le plus grand nombre de réponses correctes. 
En évaluant nos résultats de cette manière nous observons que les questions \textit{à quoi, de quand} ont des performances excellentes, ceci est dû au fait que ces questions sont fortement reliées à une préposition. Les questions les plus fréquentes sont \textit{ qui est l’agent, quoi } liées aux sujets et COD dans la syntaxe, avec des F-mesures avoisinant les $70\%$ ensuite nous avons les \textit{quand, où, qui } liées aux CCT, CCL et COD. Pour tous ceux-ci, nous avons des performances à peu près équivalentes, de l'ordre de $55\%$. Les sujets et COD sont plus faciles à détecter, car leurs chemins de dépendances sont souvent plus simples et le nombre d’exemples d’apprentissage est plus grand. Les questions pour lesquelles nous obtenons les performances les plus basses sont \textit{ dans quelle circonstance, avec quelle conséquence, de quelle manière } ce sont des questions qui ont une énorme variabilité au niveau syntaxique, sont moins fréquentes et ne sont pas ancrées à une préposition spécifique.

Dans le tableau \ref{table:FeatureSelection}, nous cherchons à mesurer l’impact de chaque caractéristique sur les performances. L’analyse est faite sur la tâche de SR. Le chemin de dépendances simplifié et la partie du discours (POS) sont les caractéristiques les plus importantes pour améliorer les performances de notre système. Par ailleurs, nous voyons que la précision est plus affectée par les lemmes, alors que le rappel est affecté par les POS, le chemin de dépendances et la distance linéaire à la cible. 
Le lemme du mot parent dans l’analyse en dépendances permet aussi d’augmenter la précision de notre système. En effet lorsque deux compléments ont des chemins de dépendances similaires (par exemple « dans le journal » et « pendant la guerre »), ils sont faciles à classer grâce à leur tête syntaxique.
La pertinence de ce paramètre  est liée au fait que l’analyse en dépendances a été faite en suivant une convention similaire à celle du French Treebank \cite{abeille2003building,abeille2004enriching} et donc en considérant les prépositions comme tête des sous-arbres. 
Pour la dernière ligne du tableau, seuls les paramètres Lemme, POS et distance linéaire sont utilisés, et nous pouvons constater une perte de 4.5 points de F-mesure par rapport au système qui se sert de l’analyse en dépendances.   

\begin{table}[h]
  \begin{center}
    \begin{tabular}{| l | c | c | c | }
      \hline
      \multicolumn{1}{ |c| }{ Paramètres } & Précision & Rappel & F-mesure  \\ \hline
      Tous les paramètres  & $82.2 \pm 0.6$ & $51.2 \pm 0.7$ & $\textbf{63.1} \pm 0.6$ \\ \hline
      Tous sauf Chemin Dépendance   & $82.5 \pm 0.8$ & $47.2 \pm 0.7$ & $\textit{60.0} \pm 0.6$ \\ \hline
      Tous sauf Partie du Discours (POS)    & $83.0 \pm 1.0$ & $47.1 \pm 1.0$ & $\textit{60.1} \pm 1.0$ \\ \hline
    Tous sauf Distance Linéaire  & $82.2 \pm 0.7$ & $48.6 \pm 0.7$ & $61.1 \pm 0.7$ \\ \hline
    Tous sauf Lemme  & $80.2 \pm 0.6$ & $50.9 \pm 0.7$ & $62.2 \pm 0.6$ \\ \hline
    Tous sauf Lemme Parent & $81.0 \pm 0.9$ & $51.0 \pm 0.8$ & $62.6 \pm 0.8$ \\ \hline
    Tous sauf Analyse en Dépendance & $80.8 \pm 1.2$ & $45.9 \pm 0.7$ & $\textit{58.6} \pm 0.7$ \\ \hline
    \end{tabular}
    \caption{Effets de l'élimination de chaque paramètre sur les performances}
    \label{table:FeatureSelection}
  \end{center}
\end{table}

\setlength{\textfloatsep}{2pt}
\subsection{Influence de la complexité des textes}
\vspace{-8pt}
Chaque phrase a une complexité inhérente, qui est due à divers facteurs. D’une façon très simpliste, une phrase plus longue est souvent plus complexe et difficile à traiter. Si nous n'observons pas d'influence sur la sélection des cadres (SC), la longueur des phrases s'avère très importante pour la tâche de SR. 
Nous avons observé en effet une perte de précision de plus de 7 points et une perte en rappel de plus de 22 points entre les phrases du premier décile (8 mots par phrase en moyenne) et les phrases du dernier décile (50 mots par phrase en moyenne), avec une décroissance monotone du rappel sur les 10 déciles. Ceci est dû au fait que les phrases très longues ont souvent plus de rôles sémantiques et des rôles sémantiques plus rares. 
De façon analogue, chaque UL a une complexité inhérente, qui dépend du fait que ce soit un verbe ou un substantif, et de la position qu’elle occupe dans l’arbre de dépendances de la phrase. Une cible est dite « racine » lorsqu’elle constitue la racine de l’arbre de dépendance de sa phrase, et « non racine » dans le cas contraire.  
\setlength{\intextsep}{8pt}
\begin{table}[h]
  \begin{center}
    \begin{tabular}{ | c | c | c | c | c | c | }
      \hline
      Type de Cible & Nb Cibles & Nb FE & Précision & Rappel & Fmesure \\ \hline
      Verbe Racine     & $5389$ & $13592$ & $\textbf{85.4} \pm 0.3$ & $\textbf{68.2} \pm 1.4$ & $\textbf{75.9} \pm 0.9$  \\ \hline
      Verbe non Racine & $8532$ & $19496$ & $83.0 \pm 0.9$ & $51.3 \pm 1.2$ & $63.4 \pm 0.8$ \\ \hline
      Nom Racine       & $279$ & $252$ & $72.2 \pm 7.6$ & $50.2 \pm 6.9$ & $59.0 \pm 6.5$ \\ \hline
      Nom non Racine   & $7198$ & $13538$ & $75.4 \pm 2.1$ & $34.2 \pm 0.8$ & $47.0 \pm 0.9$ \\ \hline
      Total            & $21398$ & $46878$ & $82.2 \pm 0.6$ & $51.2 \pm 0.7$ & $63.1 \pm 0.6 $ \\ \hline
    \end{tabular}
    \caption{Résultats de la sélection de rôles par type de cible}
    \label{table:TargetType}
  \end{center}
\end{table}
\vspace{-10pt}
En analysant le détail des performances par Unité Lexicale, on observe de grandes disparités dans les résultats, avec 8 UL qui produisent une F-mesure supérieure à $75\%$, et 8 qui conduisent à une F-mesure inférieure à $25\%$. La quantité de données d'apprentissage n'est pas le seul facteur explicatif.
Parmi les UL qui ont plus de 1000 occurrences dans le corpus, 2 UL nominales ont des performances très moyennes autour de $40\%$ ( \textit{combat} et \textit{attaque}) alors que les deux UL (\textit{provenir} et \textit{contenir}) qui produisent les meilleurs résultats (F-mesure supérieure à $80\%$) n'ont que 200 échantillons dans le corpus.  
Dans le tableau \ref{table:TargetType} nous montrons que la position de la cible dans l’arbre de dépendance a un impact important sur le rappel, avec une différence de plus de 15 points entre le cas des cibles  « racine » et « non racine ». 
Les cibles « non racine » présentent des chemins plus compliqués et moins fiables vers leurs rôles sémantiques. Lorsqu’on compare les UL nominales avec les UL verbales, il y a une différence d'environ 10 points sur la précision et d'environ 17 points pour le rappel. Même si les cibles nominales ont moins de rôles sémantiques associés (2.3 rôles sémantiques par cadre verbal contre 1.8 par cadre nominal en moyenne) elles demeurent plus complexes à traiter, car les chemins de dépendance vers leurs rôles sémantiques sont très variables. Il faut aussi prendre en compte le fait que les UL nominales sont plus rares et ont moins de données d'apprentissage associées. 

\vspace{-8pt}

\subsection{Influence des données d'apprentissage}

Vue la complexité de l’annotation manuelle des cadres sémantiques, la génération de nouvelles ressources n'est pas toujours possible. Pour extraire ces cadres sur des documents d’un nouveau domaine ou issus d'une nouvelle source, nous nous intéressons à évaluer les performances des modèles appris sur des données d’autres sources annotées. 
Comme dans cette expérience nous nous intéressons à évaluer l’impact de la similarité thématique sur les performances du système, nous avons réduit notre jeux de données aux 54 UL qui étaient présentes dans nos 4 corpus.
\begin{table}[h]
  \begin{center}
    \begin{tabular}{  c | c | c | c | c | }
	   \cline{2-5}
       & Taille App. & Précision & Rappel & Fmesure \\ \hline
      \multicolumn{1}{ |c|  }{ 80\% CTGM } & $304$ & $83.1 \pm 9.7$ & $15.1 \pm 2.0$ & $25.5 \pm 3.2$  \\ \hline
       \multicolumn{1}{ |c| }{ 80\% WA } & $3264$ & $78.6 \pm 8.1$ & $26.1 \pm 4.6$ & $39.1 \pm 5.9$ \\ \hline
      \multicolumn{1}{ |c| }{ 40\% WGM } & $2918$ & $77.1 \pm 8.4 $ & $32.2 \pm 5.4$ & $45.2 \pm 5.8$ \\ \hline
      \multicolumn{1}{ |c| }{ 80\% WGM } & $5836$ & $80.3 \pm 7.1 $ & $37.8 \pm 4.6$ & $51.3 \pm 4.8$ \\ \hline
      \multicolumn{1}{ |c|  }{ 80\% WGM + 80\% WA + 80\%VKH } & $9413$ & $78.6 \pm 7.7$ & $39.0 \pm 5.2$ & $52.0 \pm 5.7$ \\ \hline
      \multicolumn{1}{ |c|  }{ 80\% WGM + 80\% CTGM } & $6140$  & $79.8 \pm 5.4$ & $39.9 \pm 3.5$ & $53.1 \pm 3.6$ \\ \hline
      \multicolumn{1}{ |c|  }{ 80\% ALL }  & $9717$ & $79.3 \pm 5.9$ & $41.2 \pm 2.3$ & $54.1 \pm 2.7$ \\ \hline
    \end{tabular}
    \caption{Effets de la constitution du corpus d'apprentissage}
    \label{table:TrainCorpus}
  \end{center}
\end{table}
\vspace{-8pt}    
Pour cette expérience nous considérons que nos documents issus de CTGM sont une nouvelle source. Cliotexte regroupe des textes historiques (discours, déclarations, ...) qui ne correspondent pas exactement à un style encyclopédique. Nous proposons diverses répartitions du corpus d’apprentissage et nous mesurons les performances des systèmes pour chaque configuration. Dans le tableau \ref{table:TrainCorpus} nous montrons que pour une même taille de corpus d’apprentissage et un style fixe (données issues de Wikipedia, 80\%WA vs. 40\%WGM) les performances obtenues avec un corpus du même domaine thématique (40\%WGM) sont supérieures par 6 points de F-mesure. Nous arrivons à des performances moyennes rien qu’avec un apprentissage fait avec un corpus du même domaine (80\%WGM), sans avoir utilisé aucune données annotées de CTGM. De plus, à partir du moment où WGM est inclus dans le corpus d’apprentissage, l'ajout de 3500 exemples hors domaine n’a pas eu d'impact important alors que le simple ajout de 304 exemples de cadres issus du CTGM, augmente les performances de 2 points de F-mesure. Ceci met en évidence le fait qu’il est toujours utile d’annoter quelques exemples des phrases de la même source, pour franchir les différences de vocabulaire et de style.

\vspace{-18pt}

\section{Conclusion}
\vspace{-8pt}
Dans cet article nous avons présenté la tâche d’analyse en cadres sémantiques comme un problème d’étiquetage de séquences que nous avons abordé à l’aide de modèles CRF.  Nous avons effectué diverses expériences faites sur le corpus CALOR constitué de données encyclopédiques annotées en cadres sémantiques, montrant des performances encourageantes à partir de données d'apprentissage de taille moyenne. Les résultats obtenus révèlent une grande variabilité des performances en fonction des types d'unité lexicale (verbales ou nominales), des types de rôles sémantiques (relations directes ou circonstancielles) mais également en fonction de la complexité intrinsèque des phrases considérées (longueur, structure de dépendance). 
Dans nos futurs travaux, nous allons explorer des modélisations par étiquetage de séquences à l’aide de modèles neuronaux RNN, LSTM; et nous allons nous intéresser également au partage d’information entre les rôles des différents cadres sémantiques et UL, pour pouvoir mieux traiter les cas des cibles et cadres sémantiques peu fréquents.

\bibliographystyle{taln2017}
\bibliography{true_biblio}

\end{document}